**Title:** GPT-4 to GPT-3.5: 'Hold My Scalpel' – A Look at the Competency of OpenAI's GPT on the Plastic Surgery In-Service Training Exam

**Authors:** Jonathan D. Freedman MD, PhD[1], Ian A. Nappier AB

1. Division of Plastic Surgery, Department of Surgery, University of Colorado

**Correspondence:**

Jonathan Freedman, MD PhD

University of Colorado, Anschutz Medical Center

Division of Plastic and Reconstructive Surgery

12631 East 17th Ave, Aurora, CO 80045

JonathanDFreedman@gmail.com

Ian Nappier

iannappier@gmail.com

**Disclosures:**

The authors did not receive any funding for this study. None of the authors has a financial interest in any of the products, devices, or drugs mentioned in this manuscript.

# ABSTRACT/SUMMARY

The Plastic Surgery In-Service Training Exam (PSITE) is an important indicator of resident proficiency and serves as a useful benchmark for evaluating OpenAI's GPT. Unlike many of the simulated tests or practice questions shown in the GPT-4 Technical Paper, the multiple-choice questions evaluated here are authentic PSITE questions. These questions offer realistic clinical vignettes that a plastic surgeon commonly encounters in practice and scores highly correlate with passing the written boards required to become a Board Certified Plastic Surgeon. Our evaluation shows dramatic improvement of GPT-4 (without vision) over GPT-3.5 with both the 2022 and 2021 exams respectively increasing the score from 8th to 88th percentile and 3rd to 99th percentile. The final results of the 2023 PSITE are set to be released on April 11, 2023, and this is an exciting moment to continue our research with a fresh exam. Our evaluation pipeline is ready for the moment that the exam is released so long as we have access via OpenAI to the GPT-4 API. With multimodal input, we may achieve superhuman performance on the 2023.

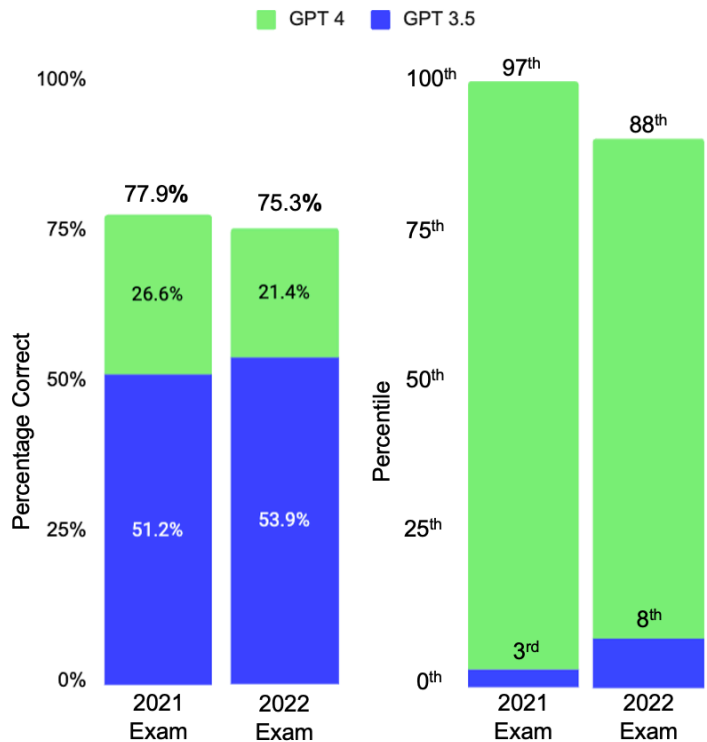

**Abstract Figure.** GPT-3.5 and GPT-4 performance on the 2021 and 2022 PSITEs. Substantial improvement percentage wise led to big increases in percentile score. This would place GPT-4's performance amongst the top echelon of plastic surgery residents.

MANUSCRIPT

**Introduction**

Training in plastic and reconstructive surgery is an extensive and rigorous apprenticeship, encompassing the mastery of microsurgery, hand and extremity surgery, craniomaxillofacial surgery, cosmetic surgery, and fundamental surgical principles. Trainees follow one of two routes: a six-year integrated program or a three-year independent program after completing general surgery or another surgical specialty. Annually, plastic surgery residents undertake the Plastic Surgery In-Service Training Examination (PSITE), administered by the American Society of Plastic Surgeons (ASPS). This exam tests their knowledge on the "most up-to-date, challenging and comprehensive" material, with five sections of 50 questions each: Comprehensive, Hand and Lower Extremities, Craniomaxillofacial, Cosmetic and Breast, and Core Surgical Principles. The results serve as an indicator of the resident's overall knowledge with minimum percentile scores required for residency progression and as a predictor of their success in the American Board of Plastic Surgery written examination, which is taken during the final year of training.

Recently, in the field of Artificial Intelligence, Large Language Models (LLMs), such as OpenAI's GPT-4, have garnered interest for their capacity to understand and generate text that closely resembles human language. Constructed using deep learning algorithms and extensive internet-based data sets, these AI models can identify patterns, context, and connections between diverse pieces of information. This allows them to offer well-informed responses and recommendations across a broad spectrum of subjects. GPT-4 has demonstrated remarkable proficiency in multiple-choice question (MCQ) exams for various professional fields, including the bar exam, USMLE practice questions and the medical knowledge self-assessment program. With this study, we aim to evaluate GPT-4's proficiency in plastic and reconstructive surgery using PSITE MCQs from the 2022 and 2021 exams.

**Methods**

Classification of PSITE Question Types

The 2022 and 2023 plastic surgery in-service exam (PSITE) multiple choice questions (MCQs) were classified into five different categories. The *Not Scored* type are questions that were not counted in scoring the examination by the American Society of Plastic Surgeons (ASPS). The *Text Only* type contained only text in both the question stem and answer. The *Supplemental Figure* type contained an additional image or diagram in addition to a question stem and answer that only contained text. The *Table Answers* type contained the answers in the form of a table. The *Embedded Answers* type contained a supplemental image or diagram where the answers were directly embedded into the image or diagram. And the *Stem Table* type contained a table in the question stem. A spreadsheet was used to record the question type for each of the 250 questions in the 2022 and 2023 examinations and the distribution counted using the *Countif* function.

Evaluation of MCQs by GPT-3.5

In order to evaluate the performance of GPT-3.5 on the PSITE, we replicated the methodology used by OpenAI to evaluate the performance of GPT-3.5 and GPT-4 on multiple-choice exams, as documented in Appendix A.2 and A.8 of the "GPT-4 Technical Report" (OpenAI, 2023).

Our evaluation consisted of Python code that would generate a prompt and submit it to the OpenAI 'gpt-3.5-turbo' API endpoint. The GPT model then provided a response containing an explanation and answer details, from which a second request was made using GPT to extract the multiple-choice answer. We compared the final letter result (e.g., "A", "B", etc.) with the true result from the exam key, and reported a correct answer when the GPT response matched.

We utilized "few-shot learning", which provides a set of examples to the model before it is prompted to provide a new response. For each question from the PSITE exam, we provided GPT-3.5 with a prompt consisting of five example PSITE questions, along with their respective choices, correct answers, and official explanations. This prompt was then followed by the PSITE exam question and its multiple-choice answer options, without the answer, and then an additional string of text that prompted the model to first explain its answer to the question.

We used the same prompt base for all of the questions in a given PSITE and each prompt had the following structure:

```
Prompt to GPT = [prompt base of 5 examples] + [PSITE question] + [request for response]
```

Our approach is consistent with the methodology and format of OpenAI's evaluation covering 20+ standardized exams, enabling direct comparison of performance of GPT-3.5 relative to humans in terms of percentile scores. This methodology is essentially similar to a recent evaluation of GPT performance on the United States Medical Licensing Examination (USMLE),

A detailed example of this methodology, and our prompt can be seen in the Appendix.

Evaluation of *Incorrect Responses* by GPT-4

At the date of this preprint, we did not have access to the GPT-4 API, and thus were unable to replicate the methodology used with GPT-3.5. In order to measure GPT-4 performance on PSITE exams, we utilized the ChatGPT interface which provides access to GPT-4.

For questions where GPT-3.5 answered incorrectly, we extracted the question text and recorded them in a spreadsheet. We then entered these questions sequentially into ChatGPT using the original question format in a single chat with a single question as a message. An estimate of GPT-4's performance was calculated by the questions GPT-4 correctly answered to the previous GPT-3.5 evaluation.

Post-processing of Exam Results

Each test displayed the same format of sections with questions 1-50 in the Comprehensive section, 51-100 in the Hand and Lower Extremity section, 101-150 in the Craniomaxillofacial section, 151-200 in the Breast and Cosmetic section and 201-250 in the Core Principles section. The overall average and section averages were calculated for each test and question type using the *Averageif* formula. The *Not Scored* questions were excluded from the averages and were not evaluated. The Norm table provided by the ASPS was used to convert the percentage correct to a percentile score for all residents overall and further broken down for first through third year Independent Residents and for first through sixth year Integrated Residents.

MELD Method

We further rely on the memorization effects Levenshtein detector (MELD) method offered by Nori et. al. ("Capabilities of GPT-4 on Medical Challenge Problem"). This method tests for the presence of an exam question in an LLM model's training data by splitting the text of a question in two, and comparing the response of the model to the second half of the question.

**Results**

Classification of PSITE Question Types

For the 2022 exam, the distribution of the scored questions was mainly text only at 81% followed Supplemental Figure at 14%. Table Answers, Embedded Answers and Stem Table types combined made up a minority of questions at 2%. See **Table 1** and **Figure 1** for a breakdown of the distribution of question types for the 2022 exam.

For the 2021 exam, the distribution of the scored questions was mainly text only at 85% followed Supplemental Figure at 10%. Table Answers, Embedded Answers and Stem Table types combined made up a minority of questions at 3%. See **Table 1** and **Figure 1** for a breakdown of the distribution of question types for the 2021 exam.

**Table 1.** Classification of question types for the 2022 and 2021 PSITE MCQs into *Not Scored, Text Only, Supplemental Figure, Table Answers, Embedded Answers* and *Stem Table*. The number of questions and relative percentage are listed.

| Type | 2022 | 2022 (%) | 2021 | 2021 (%) |
|---|---|---|---|---|
| *Not Scored* | 7 | 2.8% | 6 | 2.4% |
| *Text Only* | 203 | 81.2% | 212 | 84.8% |
| *Supplemental Figure* | 35 | 14.0% | 25 | 10.0% |
| *Table Answers* | 3 | 1.2% | 6 | 2.4% |
| *Embedded Answers* | 2 | 0.8% | 0 | 0.0% |
| *Stem Table* | 0 | 0.0% | 1 | 0.4% |

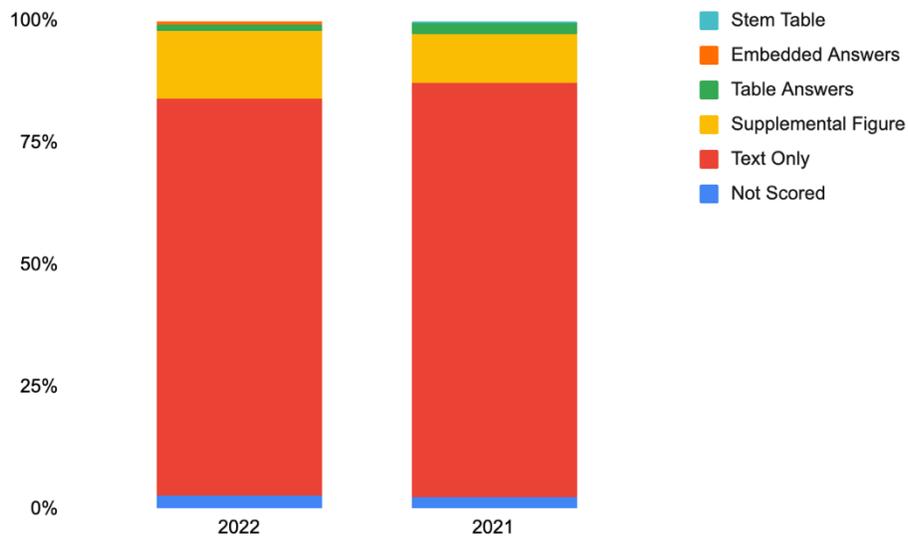

**Figure 1.** Classification of the 2022 and 2021 PSITE MCQs into *Not Scored, Text Only, Supplemental Figure, Table Answers, Embedded Answers* and *Stem Table*.

Exam Results for the 2022 PSITE

The GPT-4 estimate improved the overall percentage correct from 54% to 75%. A breakdown of GPT-3.5 and the GPT-4 performance across the 2022 exam and sections is shown in **Figure 2.** GPT-4 increased the percentage correct in each of the sections: Comprehensive increased from 55% to 76%, Hand and Lower Extremity increased from 56% to 80%, Craniomaxillofacial increased from 59% to 71%, Breast and Cosmetic increased from 43% to 68% and Core Principles increased from 57% to 82% (**Figure 2**).

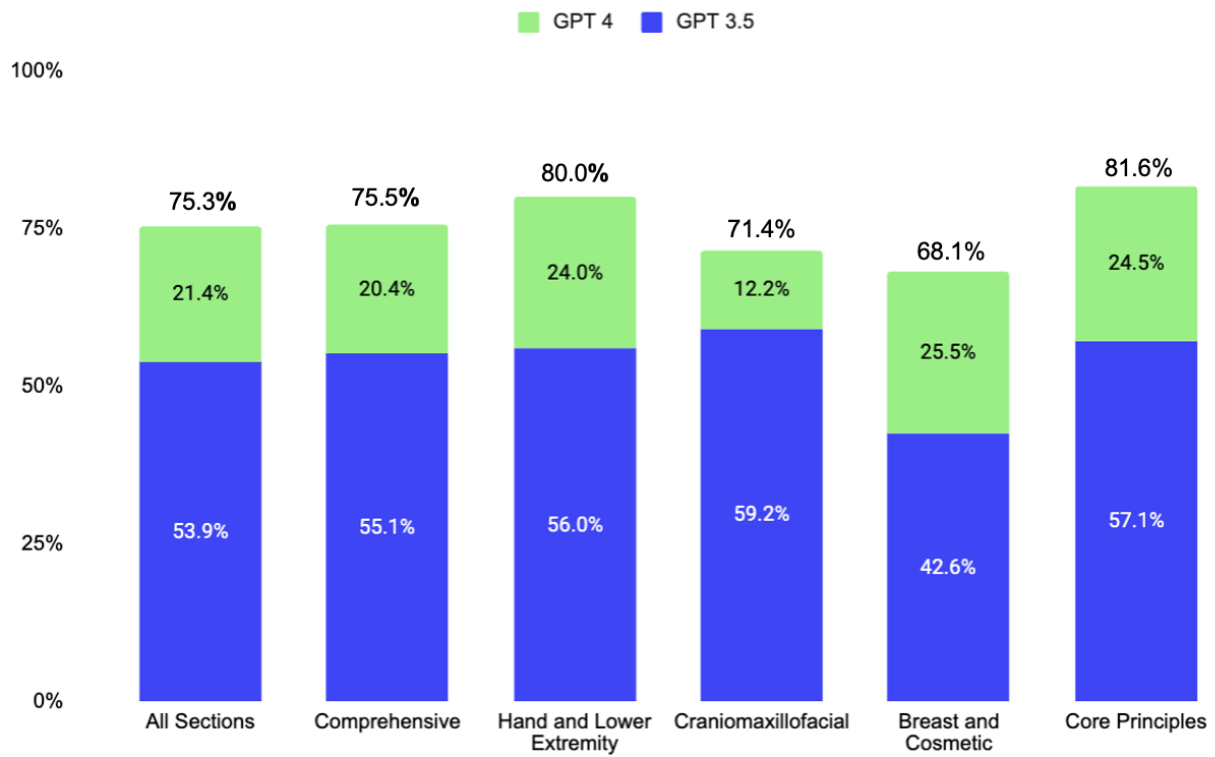

**Figure 2.** GPT performance on the 2022 PSITE. Percentage correct is shown for GPT-3.5 in blue. The increase in performance with GPT-4 is shown in green. The total percentage correct for the entire 2022 test and each section is shown at the top of each column graph.

The percentage correct was then converted to percentiles using the Norm table. GPT-4 improved the percentile score over GPT-3.5. The overall percentile score amongst all residents (n=1302) improved from the 8th percentile to the 88th percentile. The corresponding percentile score for independent and integrated residents is shown in **Figure 3**.

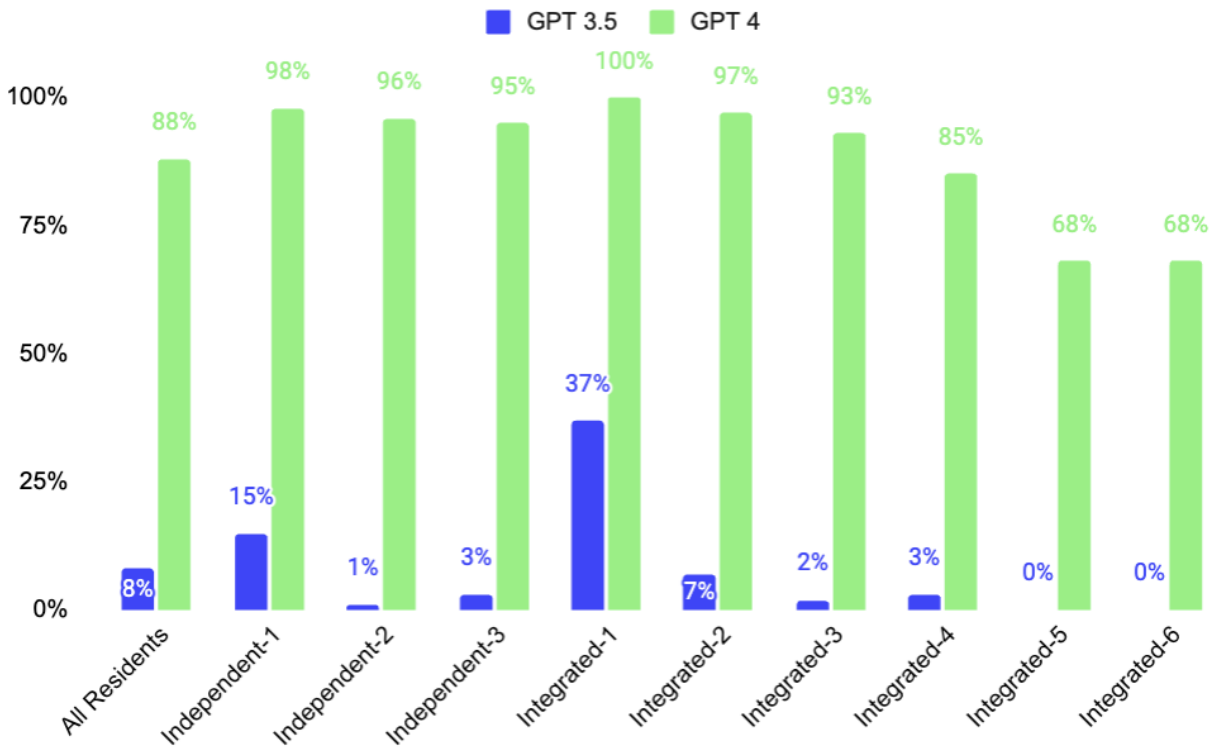

**Figure 3.** GPT-3.5 (blue) and GPT-4 (green) percentile performance among all residents and broken down between Independent years 1-3 and Integrated years 1-6.

Additionally, the percentage correct was broken down per section (Comprehensive, Hand and Lower Extremity, Craniomaxillofacial, Breast and Cosmetic and Core Principles) and per question type (*Scored, Text Only, Supplemental Figure, Table Answers* and *Embedded Answers*. *Scored, text only* and *supplemental figure* types had similar scores at 54%, 55% and 51% correct (**Figure 4**).

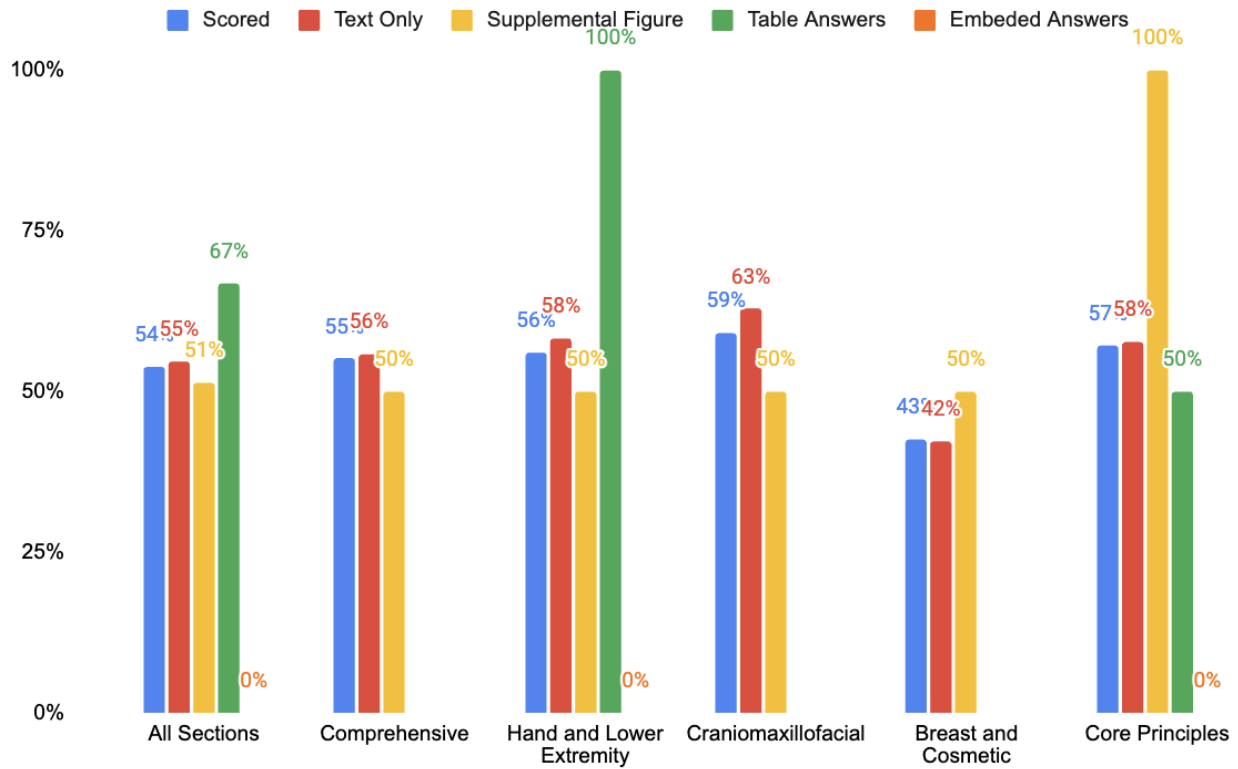

**Figure 4.** GPT-3.5 results for the 2022 PSITE broken down by question type and section.

Exam Results for the 2021 PSITE

The GPT-4 estimate improved the overall percentage correct from 51% to 78%. A breakdown of GPT-3.5 and the GPT-4 performance across the 2021 exam and sections is shown in **Figure 5.** GPT-4 increased the percentage correct in each of the sections: Comprehensive increased from 50% to 75%, Hand and Lower Extremity increased from 45% to 82%, Craniomaxillofacial increased from 63% to 69%, Breast and Cosmetic increased from 48% to 81% and Core Principles increased from 49% to 80% (**Figure 5**).

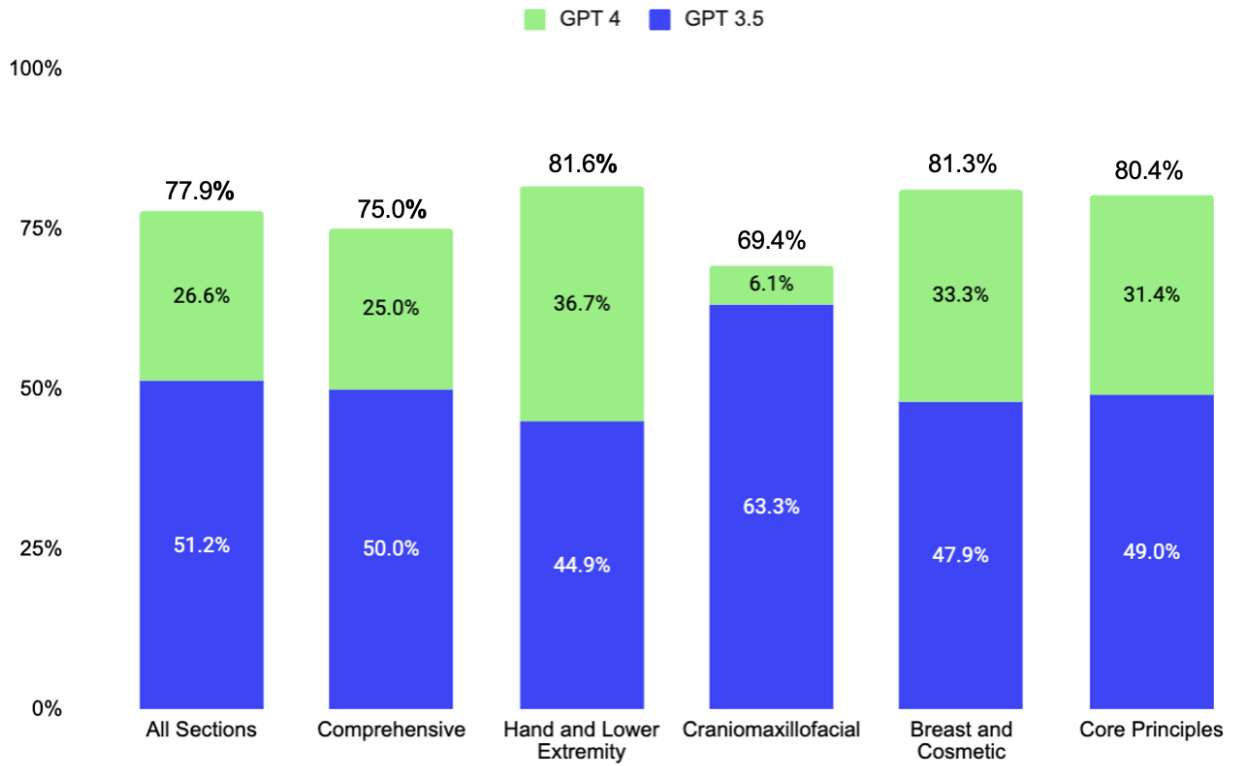

**Figure 5.** GPT performance on the 2021 PSITE. Percentage correct is shown for GPT-3.5 in blue. The increase in performance with GPT-4 is shown in green. The total percentage correct for the entire 2022 test and each section is shown at the top of each column graph.

The percentage correct was then converted to percentiles using the Norm table. GPT-4 improved the percentile score over GPT-3.5. The overall percentile score amongst all residents (n=1255) improved from the 3rd percentile to the 97th percentile. The corresponding percentile score for independent and integrated residents is shown in **Figure 6**.

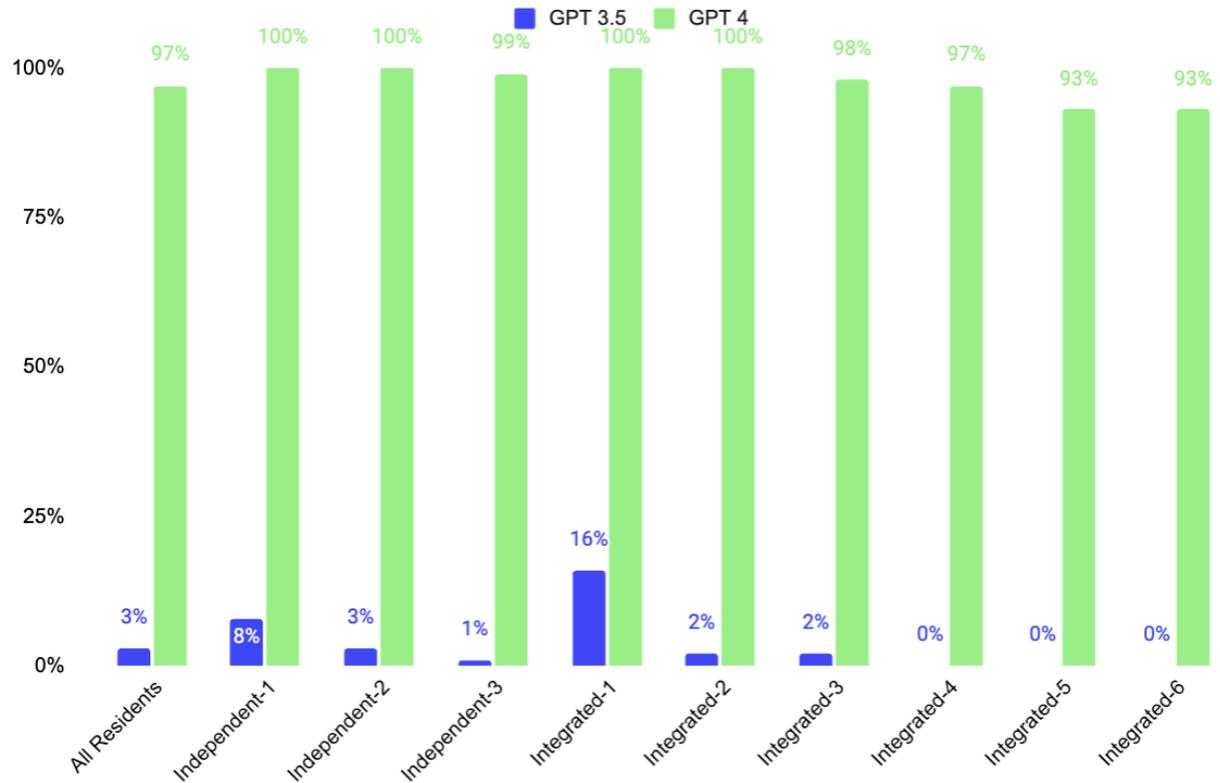

**Figure 6.** GPT-3.5 (blue) and GPT-4 (green) percentile performance among all residents and broken down between Independent years 1-3 and Integrated years 1-6.

Additionally, the percentage correct was broken down per section (Comprehensive, Hand and Lower Extremity, Craniomaxillofacial, Breast and Cosmetic and Core Principles) and per question type (*Scored, Text Only, Supplemental Figure, Table Answers* and *Embedded Answers*. *Scored, text only* and *supplemental figure* types had similar scores at 51%, 53% and 48% correct (**Figure 7**).

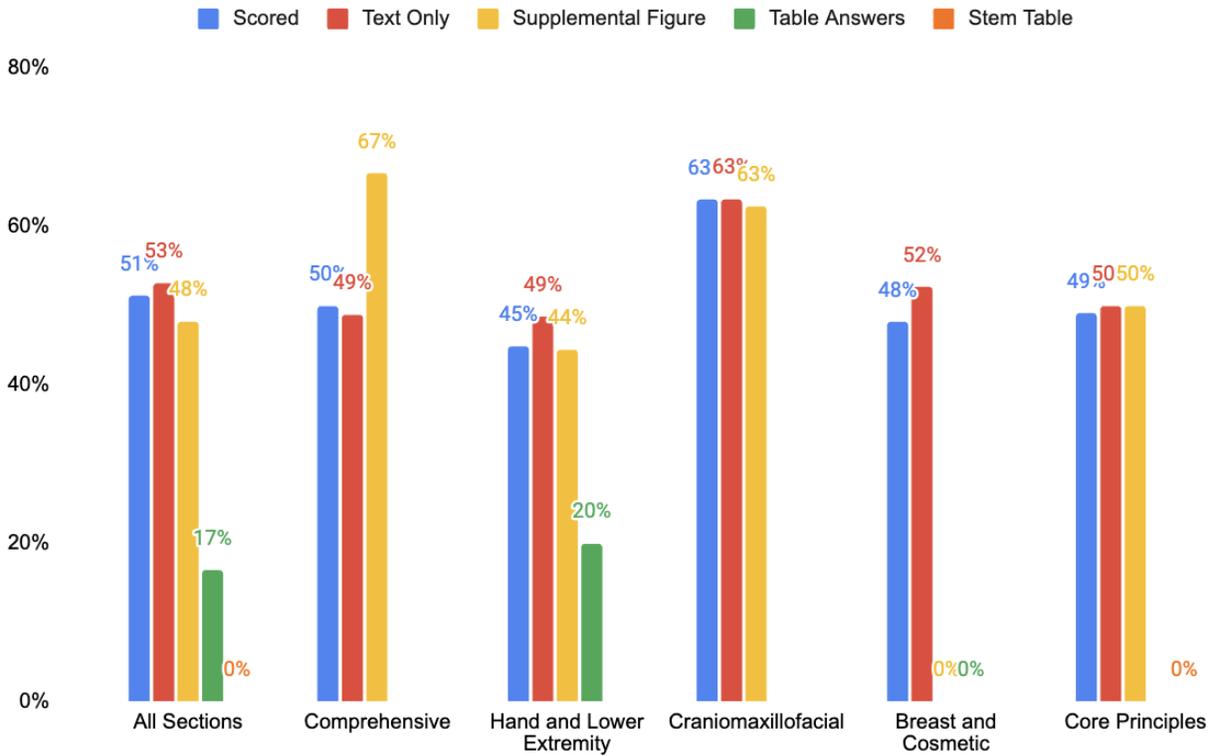

**Figure 7.** GPT-3.5 results for the 2021 PSITE broken down by question type and section.

Data Validation - MELD

Using the methodology proposed by Nori et. Al. (2023), we were unable to detect question presence in the training data of any exam question using this method with the Levenshtein distance ratio threshold suggested by Nori et. al. of 0.95 (a 95% match). Across 500 total questions evaluated using MELD, 0% of them were above the threshold value.

**Discussion**

Exam Performance

We present a novel evaluation of Open AI's GPT capabilities using the 2022 and 2021 Plastic Surgery In-Service Training Exam (PSITE) to test its proficiency. The majority of this test consists of highly relevant clinical vignettes that closely align with real-world tasks performed by plastic surgeons. These tasks include making diagnoses, ordering diagnostic tests, identifying emergency situations, selecting the best surgical approach, and prescribing medications. This test is an excellent indicator of trainee knowledge progression and serves as a valuable benchmark for evaluating GPT's performance.

GPT-4 made remarkable gains over GPT-3.5 on the 2022 and 2021 exams. On the 2022 exam, the percentage of correct questions improved from 54% to 75%. This drastically improved percentile performance from $8^{th}$ percentile to $88^{th}$ percentile for all residents. The $30^{th}$ percentile is considered passing in most programs and GPT-3.5's score would only be passing for first year integrated residents. These residents are fresh out of medical school and with limited clinical exposure they are truly at the beginning of their medical training. However, GPT-4 drastically improved performance and received a superb percentile score for all independent and integrated year residents (**Figure 3**). On the 2021 exam, again the percentage of correct questions improved from 51% to 78%. Percentile scores for the 2021 would have been considered failing for every year resident using GPT-3.5 result. Amazingly, for the GPT-4 estimate it performed above the $90^{th}$ percentile overall and for every resident year (**Figure 6**).

Data Validation and Memorization

LLMs like GPT-3.5 and GPT-4 are trained on massive datasets, much of which is scraped from the internet. A concern in an evaluation of these models is the possibility of their performance on an exam being the result of *leakage* and *memorization* due to the exam being included in the training set, and not representative of the model's true ability to perform on these exams.

To evaluate this concern we rely on two methods. The first is in examining performance across exam years that come before and after the cutoff for GPT training data, and the second is utilizing the MELD method recently offered by Nori et. al. (Nori 2023)

Training Data Date Cutoffs and Exam Performance

In this preprint, we provide results from evaluations of the 2022 and 2021 PSITE. The 2023 PSITE exam has been released, though the final answers have not been formalized.

According to the model data provided by OpenAI, the training data date cutoff for GPT-3.5 and GPT-5 is September 2021, meaning that information created after this date is not included in the training data (Open AI - Appendix). We note that 2022 PSITE was administered after this date, and that the performance of both GPT-3.5 and GPT-4 2022 and 2021 PSITE exams is consistent across these exam years (Appendix - PSITE Exam Information). The similar performance on both of these years of exams suggest that the 2021 PSITE exam was not contained within the training data for either of the GPT models evaluated. Though not finalized, initial 2023 exam questions have been released, and initial investigations show further consistency between performance on 2021 and 2022 performance.

MELD Method to detect presence in training data

To further validate that PSITE exam questions do not appear in the training data of GPT-3.5, we utilize the MELD methodology, which prompts GPT with the first half of an exam question, and compares the generated text continuing this question with the actual text of the second half of this question. According to this methodology, we did not see a positive indication that either PSITE exam was contained within the training set of GPT-3.5.

In the future, with access to the GPT-4 API, we intend on conducting further analysis and including more detailed results of this evaluation across exam years and threshold levels as we are able to conduct a more complete analysis with both GPT-3.5 and GPT-4 API access, and further assess the validity of this approach.

From Text-only to Multimodal

Our evaluation of the performance of GPT models on PSITE was conducted entirely in text. However, this is not the format of the PSITE, which involves a significant proportion of questions that involve visual elements. The 2022 and 2021 exams have 16.5% and 13.1% of questions involving visual elements, respectively. In Open AI's technical paper on GPT-4 (https://openai.com/research/gpt-4) when vision was added, there were modest gains on the GRE Quantitative (3.8%), GRE Verbal (2.4%) and LSAT (1.2%) exams and a more substantial improvement on the AMC-12 exam (25%).

GPT-4 is a multimodal model, which can simultaneously accept as input both text and image information. Without API access at time of this preprint, we were unable to use the multimodal features of GPT-4 and used it in a text-only mode. With API access, we are eager to conduct further evaluations utilizing visual input.

Given our initial results using GPT-4, and the significant percentage of these visual questions, even a small increase in performance could move overall performance into the 99th percentile compared to all surgical residents. In short, superhuman performance on the PSITE (See **Figure 3**, **Figure 6** and the Norm Tables in the Appendix).

Limitations in Comparing GPT-4 Accessed via ChatGPT to GPT-3.5

Our access to GPT-4 was limited to the ChatGPT interface, and thus we were unable to evaluate performance of GPT-4 across all PSITE exam questions. In the near future, our training evaluation pipeline is set up so that we can do a drop-in replacement of API endpoints and run evaluations using GPT-4 in both text-only and multi-modal modes, allowing GPT-4 to be compared more directly using the same methodology as we used for GPT-3.5.

Due to rate limitations of ChatGPT, we used GPT-4 only on questions where GPT-3.5 answered in*correctly*. To assess the validity of this approach, we also submitted 25 questions from the 2022 PSITE (5 from each of the 5 questions of the PSITE) that GPT-3.5 answered *correctly* using GPT-4 via ChatGPT. We saw that GPT-4 answered 22 out of 25 questions (88%) of them correctly. This suggests that the improvement in exam performance from GPT-3.5 is valid and should be representative of what we will see in future results utilizing the GPT-4 API.

API Access to GPT-4 and the 2023 PSITE

For both of the above reasons, we need access to the GPT-4 API in order to advance our research. This will allow a multimodal assessment of GPT-4 performance, a more accurate comparison with the human experience of taking the PSITE. We will also increase our confidence in the final performance of GPT-4 on the PSITE when we have full control over the parameters of our request made to the GPT-4 API, allowing us to conduct a complete evaluation of the PSITE using GPT-4 with the same methodology used for GPT-3.5

With the final results of the 2023 PSITE set to be released on April 11, 2023, this is an exciting moment to continue our research with a fresh exam. Our evaluation pipeline is ready for the moment that the exam is released so long as we have access via OpenAI to the GPT-4 API.

**Conclusion**

The Plastic Surgery In-Service Training Exam (PSITE) is an important indicator of resident proficiency and serves as a useful benchmark for evaluating OpenAI's GPT. Unlike many of the simulated tests or practice questions shown in the GPT-4 Technical Paper, the multiple-choice questions evaluated here are authentic PSITE questions. These questions offer realistic clinical vignettes that a plastic surgeon commonly encounters in practice and scores highly correlate with passing the written boards required to become a Board Certified Plastic Surgeon. Our evaluation shows dramatic improvement of GPT-4 (without vision) over GPT-3.5 with both the 2022 and 2021 exams respectively increasing the score from 8th to 88th percentile and 3rd to 99th percentile. The final results of the 2023 PSITE are set to be released on April 11, 2023, and this is an exciting moment to continue our research with a fresh exam. Our evaluation pipeline is ready for the moment that the exam is released so long as we have access via OpenAI to the GPT-4 API. With multimodal input, we may achieve superhuman performance on the 2023.

# Appendix

## Detailed Prompting Information

Our prompt base consisted of 5 example questions drawn from PSITE exams offered before the year 2017 and thus outside of the exam questions we evaluated GPT performance on. We drew 5 questions with one of each coming from the 5 standard sections of PSITE,

Each of our prompts would start with

Prompt base:

```
ANSWER KEY

Here are the answers for the problems in the exam.

Problem 1. An 8-year-old boy is brought to the emergency department
after sustaining injury to the right upper extremity, 3-cm proximal
to the antecubital fossa. Which of the following factors is
associated with improved functional outcomes
following peripheral nerve repair?
[A] Fewer suture strands used in the nerve repair
[B] Higher-tension nerve repair
[C] Increasing time between nerve injury and repair
[D] More proximal nerve injury
[E] Younger patient age

Explanation for Problem 1: The repair of peripheral nerve injuries
can be affected by several factors. Younger patients tend to have
improved outcomes compared with older patients.
Although there is no consensus on the optimal timing for nerve
repair, earlier repairs have been shown to have better outcomes than
those attempted at later time points. The level at which the injury
has occurred can also affect the outcome. The more proximal the
injury, the worse the prognosis in terms of motor and sensory return.
Moreover, more complete and rapid regain of function occurs in more
proximally innervated muscles. Finally, technical aspects of the
nerve repair can also affect outcomes. Minimal tension and an
```

increasing number of suture strands crossing the repair site are both associated with improved function.

The answer is therefore [E]

Problem 2. A 30-year-old man is brought to the emergency department because of ring avulsion of the right ring finger with complete amputation through the proximal phalanx. Problem Which of the following factors is most likely to influence survival of the replanted finger in this patient?
[A] Associated fracture of the middle phalanx [B] Level of amputation
[C] Need for skin graft
[D] Number of dorsal veins repaired/grafted
[E] Patient history of cigarette smoking

Explanation for Problem 2: The only factor that correlated with survival in the reported series and reviews is the repair and/or grafting of two or more dorsal veins. Reports that did not compare groups recommend repairing at least two dorsal veins of replanted digits.
Smoking, level of amputation, need for skin grafting, and associated fractures were not found to have any effect on survival of digits that had ring avulsion injuries.

The answer is therefore [D]

Problem 3. A 35-year-old woman undergoes surgical resection of a left parotid gland malignancy. The facial nerve was resected with the tumor, leaving a 5- to 7-cm gap between the proximal nerve stump and the distal nerve branches. Which of the following is the most appropriate treatment?
[A] Cable nerve grafting
[B] Cross-facial nerve grafting
[C] Hypoglossal nerve to facial nerve transfer
[D] Innervated gracilis muscle free flap reconstruction
[E] Nerve repair with a conduit

Explanation for Problem 3: When a facial nerve has been divided or resected, the best outcomes for regaining function are usually obtained from direct repair, or cable nerve grafting when too great a distance for direct repair separates the nerve ends.
While autologous nerve grafts from "expendable" donor nerves, such as the great auricular nerve or sural nerve, have long been the gold standard, nerve repair using biologic or synthetic nerve conduits has also produced reasonable results, in some series equivalent to cable nerve grafts. Conduit nerve repair has the advantage of having no donor site morbidity. However, the length of the gap between the proximal and distal cut nerve ends is usually limited to less than 3 cm for the best chances of nerve recovery.

When direct repair or cable nerve grafting is not possible—for example, when the nerve has been resected very proximally up to the intracranial portion of the nerve— cross-facial nerve grafting between redundant branches of the normal contralateral nerve and the distal facial nerve stumps of the paralyzed side can produce reasonable results with spontaneous symmetrical facial movement. Performing a nerve transfer from a donor nerve, such as the masseteric (V), spinal accessory (XI), or hypoglossal (XII) nerve can provide facial tone and symmetry at rest, and, in some cases, volitional movement with training. A temporary nerve transfer to these nerves is sometimes performed as a "babysitter" procedure while awaiting axonal growth through cross-facial nerve grafts.

When nerve repair or nerve transfers from the contralateral face or donor nerves are not feasible, such as after motor endplate degeneration has occurred in the facial muscles, innervated free muscle flap transfers can restore facial movement to the lower face. Muscles commonly used for facial reanimation include the gracilis, pectoralis minor, serratus anterior, and latissimus dorsi, because of their thinness, good excursion, and low donor site morbidity. In addition to a microvascular anastomosis, an epineural nerve repair is performed either to a cross-facial nerve graft or to a donor cranial nerve such as the masseteric nerve.

The answer is therefore [A]

Problem 4. A 52-year-old woman comes to the office to receive botulinum toxin type A injections to the corrugator and procerus. She returns to the office 1 week later because she is upset that her eyelids on both sides are droopy. Physical examination shows bilateral ptosis. Which of the following is the most appropriate treatment to improve this patient's condition until the effects of the botulinum toxin type A subside?
[A] Apraclonidine
[B] Artificial tears
[C] Botulinum toxin type A to the lateral orbicularis oculi
[D] Ophthalmic tobramycin and dexamethasone
[E] Tetracaine

Explanation for Problem 4: This patient has developed true eyelid ptosis from her botulinum toxin type A treatments coming into contact and affecting the levator palpebrae superioris muscle within the eyelid. Although the effects of botulinum toxin type A on any muscle are irreversible with medications, attempts to lessen the severity of the ptosis have been made with the use of eyedrops to stimulate the Müller muscle, which is located deep to the levator. Apraclonidine is an alpha-adrenergic agonist and as such stimulates the Müller muscle to contract. This contraction may elevate the eyelid 1 to 3 mm and lessen the amount of ptosis to varying degrees in order to make the overall appearance of the eyelids more tolerable to the patient until the effects of botulinum toxin type A wear off on their own and levator function naturally returns. The most common dose of apraclonidine is 1 to 2 drops three times daily until ptosis resolves.
Tetracaine is a commonly used numbing agent for the corneal surface that enhances the comfort of using corneal protectors for periorbital surgery. Tobradex eyedrops are a combination of tobramycin and dexamethasone used for treatment of infection and/or its anti-inflammatory effect in the periorbital region. It has no effect on eyelid ptosis. Artificial tears are lubricating drops and have no effect on muscular action.

The answer is therefore [A]

Problem 5. A 26-year-old woman who is at 32 weeks' gestation sustains a traumatic head injury during a boating collision. CT scan shows subarachnoid hemorrhage and pan-facial fractures. The patient is cleared by the neurosurgeon for facial fracture repair. In the ICU, blood pressure is 112/70 mmHg and heart rate is 95 bpm. Fetal monitoring shows no distress. The patient is taken to the operating room and placed in supine position. On the operating table, blood pressure is 80/50 mmHg and heart rate is 130 bpm. Which of the following is the most appropriate next step in management?
[A] Administer fluid bolus intravenously
[B] Logroll the patient to the left
[C] Obtain immediate chest x-ray study
[D] Prepare and drape the patient for the planned procedure
[E] Start vasopressors

Explanation for Problem 5: The most appropriate next step in this scenario is to logroll the patient 4 to 6 inches (or 15 degrees) to the left, decompressing the inferior vena cava (IVC). Women in the second half of pregnancy may become hypotensive when placed in the

```
supine position, caused by compression of the inferior vena cava by
the enlarged uterus, reducing venous return to the heart by up to
30%. Spinal precautions should be maintained for any patient whose
spine has not been appropriately cleared.
Vasopressors should be used as a last resort in restoring the blood
pressure of pregnant trauma patients, as these drugs further reduce
uterine blood flow, resulting in fetal hypoxia. The placental
vasculature is exquisitely sensitive to catecholamine stimulation.
Crystalloid fluid resuscitation would be indicated if the patient's
vital signs did not return to baseline after repositioning and IVC
decompression. Similarly, a chest X- ray could be obtained as part of
the workup for unresponsive hypotension.
Ignoring this patient's hemodynamic changes and proceeding with
surgery would be a mistake, as the placenta would most likely be
deprived of vital perfusion, resulting in fetal distress.

The answer is therefore [B]

Problem 6.
```

Using this prompt base, we would then concatenate with the text of the question we were evaluating from a PSITE exam and send this to GPT with the following parameters, replicating OpenAI methodology (OpenAI 2023)

```
<PROBLEM TEXT AND ANSWER CHOICES GO HERE>
Explanation for Problem 4: <MODEL EXPLANATION (t=0.3, n=1,
max_tokens=512,
stop='\nThe answer is therefore') SAMPLED HERE>
```

Thes result of this call to the GPT API is a a response to the question that contains some form of explanation and answer.

Following this, we would addend the text below, and prompt again, making a second API request in order to extract the answer letter to the PSITE question:

```
The answer is therefore [<MODEL ANSWER CHOICE (t=0.0, n=1, stop=']')
SAMPLED
HERE>]
```

For both of these calls made to the GPT, the parameters `(t=0.3, n=1, max_tokens=512)` and `(t=0.0, n=1, stop=']')` were selected to match exactly OpenAI exam evaluation methodology (OpenAI 2023), along with replicating the style and formats of the their prompts.

# Model Information

| Model Name | Model API endpoint | Training Cutoff Date |
|---|---|---|
| GPT-3.5 | gpt-3.5-turbo | September 2021 |
| GPT-4 | gpt-4 | September 2021 |

# PSITE Exam Information

| Exam Year | Exam Administration Date | Exam Release Date |
|---|---|---|
| 2021 | March 2021 | April 2021 |
| 2022 | March 2022 | April 2022 |
| 2023 | March 2023 | April 2023 |

2022 Norm Table

## 2022 NORM TABLE

| Total Test % Correct Score | Independent Program | | | | Integrated Program | | | | | | Combined Tracks | | |
|---|---|---|---|---|---|---|---|---|---|---|---|---|---|
| | All Residents (N=1,302) | First Year (N=66) | Second Year (N=70) | Third Year (N=73) | First Year (N=185) | Second Year (N=184) | Third Year (N=203) | Fourth Year (N=191) | Fifth Year (N=177) | Sixth Year (N=151) | Int 4 Ind 1 (N=257) | Int 5 Ind 2 (N=247) | Int 6 Ind 3 (N=224) |
| 86 | 100 | 100 | 100 | 100 | 100 | 100 | 100 | 100 | 100 | 100 | 100 | 100 | 100 |
| 85 | 100 | 100 | 100 | 100 | 100 | 100 | 100 | 99 | 99 | 100 | 100 | 100 | 100 |
| 84 | 100 | 100 | 100 | 100 | 100 | 100 | 100 | 99 | 98 | 99 | 100 | 99 | 100 |
| 83 | 99 | 100 | 100 | 100 | 100 | 100 | 100 | 98 | 98 | 99 | 99 | 99 | 100 |
| 82 | 99 | 100 | 100 | 100 | 100 | 100 | 100 | 98 | 98 | 99 | 99 | 99 | 100 |
| 81 | 99 | 100 | 100 | 100 | 100 | 100 | 99 | 98 | 97 | 99 | 99 | 98 | 99 |
| 80 | 98 | 100 | 100 | 100 | 100 | 99 | 99 | 97 | 96 | 94 | 98 | 97 | 96 |
| 79 | 97 | 98 | 99 | 99 | 100 | 99 | 99 | 96 | 93 | 90 | 97 | 95 | 93 |
| 78 | 96 | 98 | 99 | 99 | 100 | 98 | 99 | 96 | 90 | 86 | 96 | 93 | 90 |
| 77 | 93 | 98 | 99 | 96 | 100 | 98 | 98 | 93 | 82 | 77 | 94 | 87 | 83 |
| 76 | 90 | 98 | 99 | 96 | 100 | 98 | 95 | 86 | 75 | 75 | 89 | 82 | 82 |
| 75 | 88 | 98 | 96 | 95 | 100 | 97 | 93 | 85 | 68 | 68 | 88 | 76 | 76 |
| 74 | 82 | 98 | 94 | 89 | 100 | 96 | 90 | 77 | 58 | 50 | 82 | 68 | 63 |
| 73 | 78 | 97 | 93 | 85 | 99 | 93 | 87 | 70 | 51 | 44 | 77 | 63 | 58 |
| 72 | 73 | 97 | 89 | 82 | 98 | 91 | 79 | 63 | 43 | 36 | 72 | 56 | 51 |
| 71 | 68 | 97 | 86 | 75 | 98 | 88 | 72 | 55 | 37 | 30 | 66 | 51 | 45 |
| 70 | 61 | 94 | 76 | 67 | 96 | 82 | 64 | 49 | 27 | 24 | 60 | 41 | 38 |
| 69 | 56 | 92 | 67 | 60 | 94 | 78 | 57 | 42 | 21 | 18 | 55 | 34 | 32 |
| 68 | 52 | 89 | 67 | 48 | 93 | 76 | 52 | 34 | 17 | 15 | 48 | 31 | 25 |
| 67 | 45 | 85 | 57 | 37 | 91 | 70 | 44 | 26 | 10 | 9 | 41 | 23 | 18 |
| 66 | 41 | 80 | 56 | 34 | 89 | 60 | 39 | 24 | 6 | 8 | 38 | 20 | 17 |
| 65 | 36 | 79 | 44 | 26 | 86 | 54 | 33 | 16 | 3 | 6 | 32 | 15 | 13 |
| 64 | 33 | 68 | 39 | 25 | 84 | 49 | 29 | 13 | 3 | 5 | 27 | 13 | 11 |
| 63 | 29 | 67 | 27 | 21 | 81 | 39 | 24 | 10 | 2 | 3 | 25 | 9 | 8 |
| 62 | 25 | 56 | 23 | 16 | 74 | 34 | 20 | 8 | 1 | 2 | 21 | 7 | 7 |
| 61 | 23 | 50 | 23 | 15 | 71 | 29 | 16 | 7 | 0 | 1 | 18 | 6 | 6 |
| 60 | 19 | 42 | 19 | 12 | 64 | 25 | 11 | 5 | 0 | 0 | 14 | 5 | 4 |
| 59 | 17 | 33 | 13 | 11 | 57 | 23 | 11 | 4 | 0 | 0 | 11 | 4 | 4 |
| 58 | 14 | 29 | 11 | 10 | 52 | 16 | 7 | 4 | 0 | 0 | 10 | 3 | 3 |
| 57 | 12 | 24 | 7 | 10 | 49 | 13 | 5 | 4 | 0 | 0 | 9 | 2 | 3 |
| 56 | 10 | 18 | 4 | 7 | 45 | 8 | 4 | 3 | 0 | 0 | 7 | 1 | 2 |
| 55 | 9 | 15 | 3 | 4 | 40 | 7 | 2 | 3 | 0 | 0 | 6 | 1 | 1 |
| 54 | 8 | 15 | 1 | 3 | 37 | 7 | 2 | 3 | 0 | 0 | 6 | 0 | 1 |
| 53 | 6 | 11 | 1 | 1 | 26 | 5 | 2 | 2 | 0 | 0 | 4 | 0 | 0 |
| 52 | 4 | 8 | 0 | 1 | 19 | 4 | 1 | 2 | 0 | 0 | 3 | 0 | 0 |
| 51 | 4 | 6 | 0 | 1 | 17 | 3 | 0 | 2 | 0 | 0 | 3 | 0 | 0 |
| 50 | 3 | 5 | 0 | 1 | 15 | 3 | 0 | 1 | 0 | 0 | 2 | 0 | 0 |
| 49 | 2 | 2 | 0 | 1 | 11 | 2 | 0 | 1 | 0 | 0 | 1 | 0 | 0 |
| 48 | 1 | 2 | 0 | 0 | 8 | 1 | 0 | 1 | 0 | 0 | 1 | 0 | 0 |
| 47 | 1 | 2 | 0 | 0 | 6 | 1 | 0 | 0 | 0 | 0 | 0 | 0 | 0 |
| 46 | 1 | 0 | 0 | 0 | 6 | 0 | 0 | 0 | 0 | 0 | 0 | 0 | 0 |
| 45 | 1 | 0 | 0 | 0 | 4 | 0 | 0 | 0 | 0 | 0 | 0 | 0 | 0 |
| 44 | 1 | 0 | 0 | 0 | 4 | 0 | 0 | 0 | 0 | 0 | 0 | 0 | 0 |
| 43 | 1 | 0 | 0 | 0 | 4 | 0 | 0 | 0 | 0 | 0 | 0 | 0 | 0 |
| 42 | 0 | 0 | 0 | 0 | 3 | 0 | 0 | 0 | 0 | 0 | 0 | 0 | 0 |

## 2021 Norm Table

## 2021 NORM TABLE

| Total Test % Correct Score | Independent Program[1] | | | | Integrated Program[1] | | | | | | Combined Tracks[1] | | |
|---|---|---|---|---|---|---|---|---|---|---|---|---|---|
| | All Residents (N=1,255) | First Year (N=71) | Second Year (N=74) | Third Year (N=76) | First Year (N=169) | Second Year (N=182) | Third Year (N=199) | Fourth Year (N=182) | Fifth Year (N=162) | Sixth Year (N=139) | Int 4 Ind 1 (N=253) | Int 5 Ind 2 (N=236) | Int 6 Ind 3 (N=215) |
| 83 | 100 | 100 | 100 | 100 | 100 | 100 | 100 | 100 | 100 | 100 | 100 | 100 | 100 |
| 82 | 100 | 100 | 100 | 100 | 100 | 100 | 99 | 100 | 99 | 99 | 100 | 99 | 100 |
| 81 | 99 | 100 | 100 | 100 | 100 | 100 | 99 | 99 | 98 | 98 | 100 | 99 | 99 |
| 80 | 99 | 100 | 100 | 100 | 100 | 100 | 99 | 98 | 96 | 96 | 98 | 97 | 98 |
| 79 | 98 | 100 | 100 | 99 | 100 | 100 | 99 | 97 | 95 | 95 | 98 | 97 | 96 |
| 78 | 97 | 100 | 100 | 99 | 100 | 100 | 98 | 97 | 93 | 93 | 98 | 95 | 95 |
| 77 | 96 | 100 | 99 | 99 | 100 | 99 | 96 | 95 | 90 | 91 | 96 | 93 | 93 |
| 76 | 95 | 100 | 99 | 96 | 99 | 99 | 95 | 92 | 88 | 86 | 94 | 92 | 90 |
| 75 | 91 | 100 | 96 | 93 | 98 | 98 | 93 | 85 | 79 | 81 | 89 | 84 | 86 |
| 74 | 89 | 100 | 96 | 92 | 98 | 98 | 91 | 82 | 76 | 73 | 87 | 82 | 80 |
| 73 | 84 | 100 | 93 | 88 | 98 | 98 | 88 | 74 | 67 | 64 | 81 | 75 | 73 |
| 72 | 80 | 99 | 89 | 79 | 98 | 96 | 86 | 69 | 59 | 59 | 77 | 69 | 66 |
| 71 | 77 | 97 | 89 | 78 | 98 | 95 | 83 | 63 | 53 | 52 | 73 | 64 | 61 |
| 70 | 69 | 96 | 84 | 61 | 98 | 92 | 75 | 53 | 36 | 39 | 65 | 51 | 47 |
| 69 | 65 | 93 | 84 | 57 | 96 | 89 | 68 | 48 | 31 | 31 | 60 | 48 | 40 |
| 68 | 58 | 93 | 78 | 46 | 93 | 84 | 60 | 40 | 25 | 21 | 55 | 42 | 30 |
| 67 | 55 | 90 | 72 | 43 | 92 | 80 | 59 | 36 | 21 | 15 | 51 | 37 | 25 |
| 66 | 48 | 87 | 66 | 41 | 88 | 75 | 46 | 29 | 14 | 12 | 45 | 30 | 22 |
| 65 | 45 | 86 | 59 | 36 | 87 | 66 | 41 | 25 | 12 | 9 | 42 | 27 | 19 |
| 64 | 38 | 80 | 45 | 24 | 83 | 59 | 32 | 17 | 9 | 7 | 35 | 20 | 13 |
| 63 | 34 | 73 | 42 | 16 | 81 | 53 | 28 | 11 | 7 | 6 | 28 | 18 | 9 |
| 62 | 30 | 63 | 41 | 13 | 77 | 49 | 23 | 9 | 6 | 4 | 24 | 17 | 7 |
| 61 | 26 | 56 | 31 | 11 | 72 | 42 | 16 | 7 | 4 | 3 | 21 | 12 | 6 |
| 60 | 23 | 49 | 26 | 7 | 67 | 36 | 14 | 6 | 2 | 3 | 18 | 10 | 4 |
| 59 | 18 | 42 | 19 | 5 | 59 | 25 | 10 | 4 | 2 | 1 | 15 | 7 | 2 |
| 58 | 15 | 32 | 15 | 4 | 54 | 19 | 7 | 3 | 1 | 1 | 11 | 6 | 2 |
| 57 | 12 | 25 | 9 | 1 | 44 | 16 | 6 | 3 | 0 | 1 | 9 | 3 | 1 |
| 56 | 10 | 20 | 8 | 1 | 41 | 14 | 5 | 2 | 0 | 0 | 7 | 3 | 0 |
| 55 | 8 | 14 | 4 | 1 | 34 | 11 | 5 | 2 | 0 | 0 | 5 | 1 | 0 |
| 54 | 7 | 14 | 4 | 1 | 31 | 7 | 4 | 1 | 0 | 0 | 5 | 1 | 0 |
| 53 | 6 | 13 | 3 | 1 | 25 | 5 | 3 | 1 | 0 | 0 | 4 | 1 | 0 |
| 52 | 4 | 8 | 3 | 1 | 18 | 3 | 2 | 0 | 0 | 0 | 2 | 1 | 0 |
| 51 | 3 | 8 | 3 | 1 | 16 | 2 | 2 | 0 | 0 | 0 | 2 | 1 | 0 |
| 50 | 3 | 8 | 1 | 1 | 12 | 1 | 2 | 0 | 0 | 0 | 2 | 0 | 0 |
| 49 | 2 | 6 | 1 | 0 | 10 | 1 | 2 | 0 | 0 | 0 | 2 | 0 | 0 |
| 48 | 1 | 1 | 0 | 0 | 8 | 1 | 2 | 0 | 0 | 0 | 0 | 0 | 0 |
| 47 | 1 | 1 | 0 | 0 | 5 | 1 | 1 | 0 | 0 | 0 | 0 | 0 | 0 |
| 46 | 1 | 1 | 0 | 0 | 4 | 1 | 1 | 0 | 0 | 0 | 0 | 0 | 0 |
| 45 | 0 | 0 | 0 | 0 | 1 | 1 | 1 | 0 | 0 | 0 | 0 | 0 | 0 |
| 44 | 0 | 0 | 0 | 0 | 1 | 0 | 0 | 0 | 0 | 0 | 0 | 0 | 0 |
| 43 | 0 | 0 | 0 | 0 | 1 | 0 | 0 | 0 | 0 | 0 | 0 | 0 | 0 |
| 42 | 0 | 0 | 0 | 0 | 1 | 0 | 0 | 0 | 0 | 0 | 0 | 0 | 0 |
| 41 and below | 0 | 0 | 0 | 0 | 0 | 0 | 0 | 0 | 0 | 0 | 0 | 0 | 0 |